\pdfoutput=1

\documentclass[11pt]{article}

\usepackage[final]{acl}

\usepackage{times}
\usepackage{latexsym}

\usepackage[T1]{fontenc}

\usepackage[utf8]{inputenc}

\usepackage{microtype}

\usepackage{inconsolata}

\usepackage{graphicx}

\usepackage[utf8]{inputenc}
\graphicspath{{}}

\usepackage{bm}
\usepackage{amsmath}
\usepackage{amssymb}

\usepackage{bbding}

\usepackage{booktabs}
\usepackage{multicol,multirow}
\usepackage{makecell}

\definecolor{deepred}{rgb}{0.698,0.133,0.133}
\definecolor{blue}{rgb}{0,0,1}

\usepackage{tikz}
\usepackage[edges]{forest}
\usepackage{pgfplots}
\usepackage{pifont}
\newcommand{\cmark}{\ding{51}}%
\newcommand{\xmark}{\ding{55}}%

\usepackage{array}
\usepackage{xcolor}

\usepackage[edges]{forest}
\usepackage[framemethod=tikz]{mdframed}
\usepackage{subcaption}
\definecolor{lgreen}{rgb}{0.89,0.94,0.85}
\definecolor{lred}{rgb}{0.98, 0.90, 0.84}
\definecolor{lyellow}{rgb}{1.00, 0.95, 0.80}
\definecolor{lblue}{rgb}{0.85, 0.89, 0.95}
\definecolor{hidden-draw}{RGB}{20,68,106}
\definecolor{hidden-pink}{RGB}{255,245,247}

\tikzset{%
    parent/.style =          {align=center,text width=0.7cm, rounded corners=2pt, line width=0.8mm, fill=white!0, draw=white!90},
    child/.style =           {align=center,text width=1.4cm,rounded corners=2pt, fill=blue!10,draw=blue!90,line width=0.3mm},
    T1/.style =           {align=center,text width=1.8cm,rounded corners=3pt, fill=lblue!100, draw=black,line width=0.2mm},   
    T1_end/.style =           {align=left, text width=5cm,rounded corners=5pt, fill=lblue!100,draw=blue!0,line width=0.3mm},
    T2/.style =           {align=center,text width=1.8cm,rounded corners=3pt, fill=lred!100, draw=black,line width=0.2mm},   
    T2_end/.style =           {align=left, text width=5cm,rounded corners=5pt, fill=lred!100,draw=blue!0,line width=0.3mm},
    T3/.style =           {align=center,text width=1.8cm,rounded corners=3pt, fill=lyellow!100, draw=black,line width=0.2mm},   
    T3_end/.style =           {align=left, text width=5cm,rounded corners=5pt, fill=lyellow!100,draw=blue!0,line width=0.3mm},
    T4/.style =           {align=center,text width=1.8cm,rounded corners=3pt, fill=lgreen!100, draw=black,line width=0.2mm},   
    T4_end/.style =           {align=left, text width=5cm,rounded corners=5pt, fill=lgreen!100,draw=blue!0,line width=0.3mm}
}

\title{From Seconds to Hours: Reviewing MultiModal Large Language Models on Comprehensive Long Video Understanding}

\author{Heqing Zou\textsuperscript{\ddag\S}\footnotemark[1]\footnotemark[2], Tianze Luo\textsuperscript{\ddag}\footnotemark[1],  Guiyang Xie\textsuperscript{\ddag}, Victor (Xiao Jie) Zhang\textsuperscript{\ddag}, Fengmao Lv\textsuperscript{$\flat$}, \\ \textbf{Guangcong Wang\textsuperscript{$\natural$}, Junyang Chen\textsuperscript{\pounds},  
Zhuochen Wang\textsuperscript{\ddag}, 
Hansheng Zhang\textsuperscript{\ddag}, Huaijian Zhang\textsuperscript{\ddag}} \\
\textsuperscript{\ddag}TikTok  \textsuperscript{\S}Nanyang Technological University \textsuperscript{$\flat$}Southwest Jiaotong University \\ \textsuperscript{$\natural$}Great Bay University \textsuperscript{\pounds}Shenzhen University \\
  }

\begin{document}
\maketitle
\renewcommand{\thefootnote}{\fnsymbol{footnote}}
\footnotetext[2]{This work was done when Heqing Zou was interning at TikTok, Singapore.}
\footnotetext[1]{Equal contributions.}
\renewcommand{\thefootnote}{\arabic{footnote}}
\begin{abstract}
The integration of Large Language Models (LLMs) with visual encoders has recently shown promising performance in visual understanding tasks, leveraging their inherent capability to comprehend and generate human-like text for visual reasoning. Given the diverse nature of visual data, MultiModal Large Language Models (MM-LLMs) exhibit variations in model designing and training for understanding images, short videos, and long videos. Our paper focuses on the substantial differences and unique challenges posed by long video understanding compared to static image and short video understanding. Unlike static images, short videos encompass sequential frames with both spatial and within-event temporal information, while long videos consist of multiple events with between-event and long-term temporal information. In this survey, we aim to trace and summarize the advancements of MM-LLMs from image understanding to long video understanding. We review the differences among various visual understanding tasks and highlight the challenges in long video understanding, including more fine-grained spatiotemporal details, dynamic events, and long-term dependencies. We then provide a detailed summary of the advancements in MM-LLMs in terms of model design and training methodologies for understanding long videos. Finally, we compare the performance of existing MM-LLMs on video understanding benchmarks of various lengths and discuss potential future directions for MM-LLMs in long video understanding.

\end{abstract}


\section{Introduction}
Large Language Models have demonstrated remarkable versatility and capability in understanding and generating human-like text by scaling model size and training data \cite{raffel2020exploring, brown2020language, chowdhery2023palm, touvron2023llama}. To extend these capabilities to visual understanding tasks, various methods have been proposed to integrate LLMs with specific visual modality encoders, thereby endowing LLMs with visual perception abilities \cite{alayrac2022flamingo, li2023blip}. Single images or multiple frames are encoded as visual tokens and integrated with textual tokens to help MM-LLMs achieve visual understanding. For long-video understanding, MM-LLMs \cite{dai2023instructblip, liu2024visual} are designed to process a larger number of visual frames and diverse events, enabling a wide range of real-world applications such as automatically analyzing highlight reels from sports videos, movies, surveillance footage, and egocentric videos in embodied AI.
For example, a robot could learn to make a cup of coffee from a long egocentric video. It needs to analyze key events from the long video including: 1) measuring one to two tablespoons of ground coffee for every 6 ounces of water; 2) adding water to the coffee maker's reservoir; 3) putting the coffee grounds into the filter basket; and 4) starting the coffee maker and waiting for it to brew. 
%
%
%
Modeling long-form videos with complex spatiotemporal details and dependencies remains a challenging problem \cite{wang2023selective, mangalam2024egoschema, xu2024slowfast, wu2024longvideobench}.

\begin{figure}[t]
\centering
\includegraphics[width=1.0\linewidth]{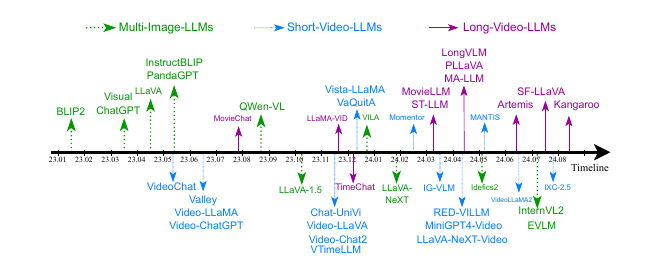}
  \caption{The development of MM-LMMs for multiple images, short videos and long videos.}
  \label{fig:time}
\end{figure}

\begin{figure*}[t]  
\centering
\includegraphics[width=1.\linewidth]{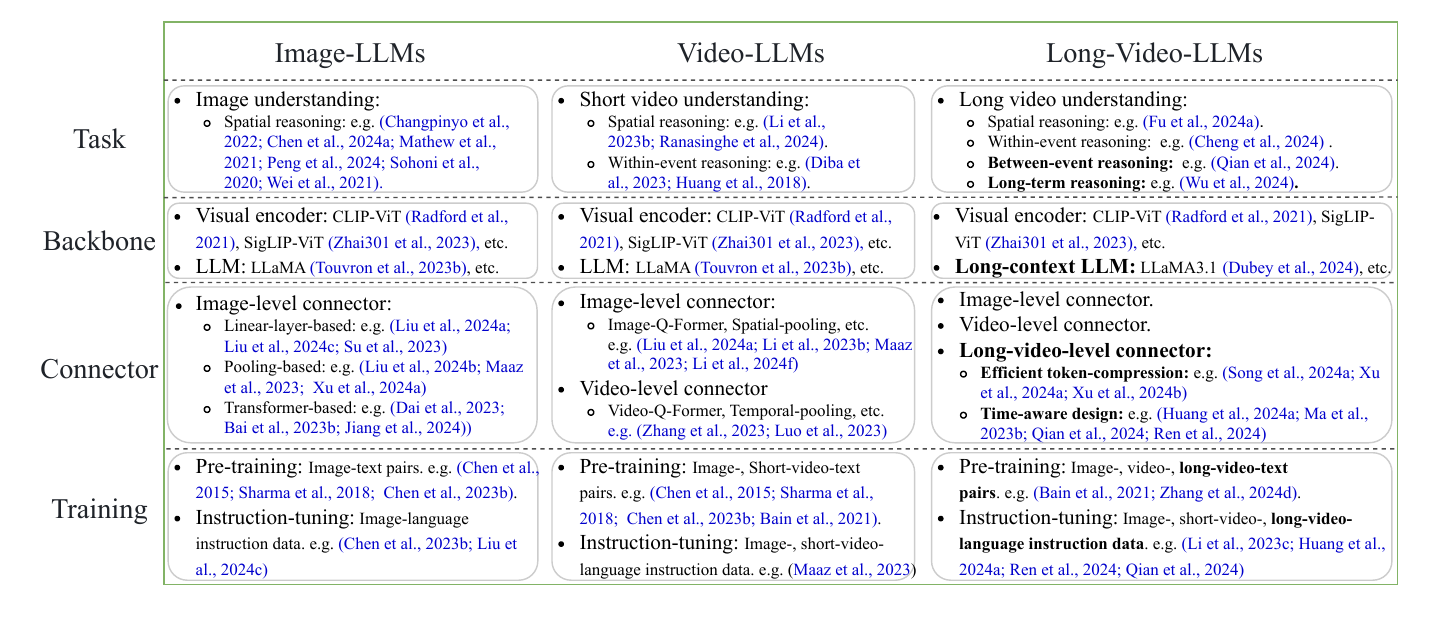}
  \caption{The comparison of MM-LLMs among Image-, Short-Video-, and Long-Video-LLMs. The \textbf{bold content} often highlights special considerations of LV-LLMs for long video understanding.}
  \label{fig:models}
\end{figure*}

There are substantial differences between long video understanding and other visual understanding tasks. Compared to static image understanding, which focuses solely on the spatial content of static images, short video understanding must also account for within-event temporal information across sequential frame changes \cite{li2023videochat, zhang2023video, maaz2023video}. Further, long videos exceeding one minute \cite{wu2021towards, zhang2024long, song2024moviechat} typically consist of multiple events with varying scenes and visual content, necessitating the capture of significant between-event and long-term variations for effective understanding. Effectively balancing spatial and temporal details with a limited number of visual tokens presents a considerable challenge for Long-Video-LLMs (LV-LLMs) \cite{song2024moviechat, he2024ma, xu2024slowfast}. Additionally, unlike short videos that span only a few seconds and contain tens of visual frames, long videos often encompass thousands of frames \cite{ren2024timechat, zhang2024long}. Therefore, LV-LLMs must be capable of memorizing and continually learning long-term correlations in videos that span minutes or even hours. The advancements of MM-LLMs \cite{fu2024video, wu2024longvideobench} for comprehensive long video understanding, especially in model design and training, warrant special attention. 

 

We summarize the comparison of MM-LLMs among Image-, Short-Video-, and LV-LLMs in Fig. \ref{fig:models}. 
In addition to the inheritance and development relationships between long-video understanding and other visual understanding tasks discussed above. LV-LLMs have also been developed by building upon the advancements of multi-image and short-video MM-LLMs, sharing a similar structure of visual encoder, LLM backbone, and cross-modality connectors. To effectively address the newly introduced challenges in long video understanding tasks, LV-LLMs are designed with more efficient long-video-level connectors that not only bridge cross-modal representations but also compress visual tokens to a manageable number \cite{li2023llama, zhang2024long}. Additionally, time-aware modules are often incorporated to enhance the capture of temporal information \cite{qian2024momentor} in LV-LLMs. For pre-training and instruction-tuning, video-text pairs and video-instruction data are essential for MM-LLMs to handle both images and videos with shared spatial perception and reasoning capacity \cite{li2023videochat}. Long video training datasets are particularly beneficial for temporal cross-modal semantic alignment and capturing long-term correlations, which are crucial for LV-LLMs \cite{song2024moviellm}. Our survey will provide a comprehensive summary of recent advances in model designing and training methods, tracing the evolution of MM-LLMs from for images to for long videos.


Recent surveys on visual understanding tasks typically adopt a single perspective, either from a global view of reviewing MM-LLMs \cite{yin2023survey, zhang-etal-2024-mm} or from a local view focusing on image- or video-understanding tasks \cite{zhang2024vision, nguyen-etal-2024-video}. While these works offer extensive reviews on the research topics, they do not discuss the developmental and inheritance relationships between different tasks and methods. Moreover, existing reviews on video understanding tasks \cite{tang2023video} tend to focus more on general video understanding rather than the more challenging task of long video understanding. Long videos over one minute are used in education, entertainment, transportation, etc., necessitating comprehensive automatic understanding with powerful models. \cite{apostolidis2021video}. Our work is among the earliest to summarize and discuss the long video understanding task from a developmental perspective.

Our survey is structured as follows: firstly, we find that the long video understanding task is more complex compared with image and short video understanding tasks (Sec.\ref{sec:2.1}), and summarize the unique challenges of long video understanding in Sec.\ref{sec:2.2}. Next, we provide a detailed summary of the developments in MM-LLMs from the perspectives of model architecture (Sec.\ref{sec:3}) and training methodologies (Sec.\ref{sec:4}), with an emphasis on the implementation of LV-LLMs for comprehensive long video understanding. We then compare the performance of video LLMs on video understanding benchmarks, from seconds to minutes (Sec.\ref{sec:5.1}) and from minutes to hours (Sec.\ref{sec:5.2}), respectively, offering insights into the existing research results of LV-LLMs. Finally, we discuss future research directions in long video understanding to advance the research field in Sec.\ref{sec:6}.

\begin{figure*}[t]
  \centering
\includegraphics[width=0.85\linewidth]{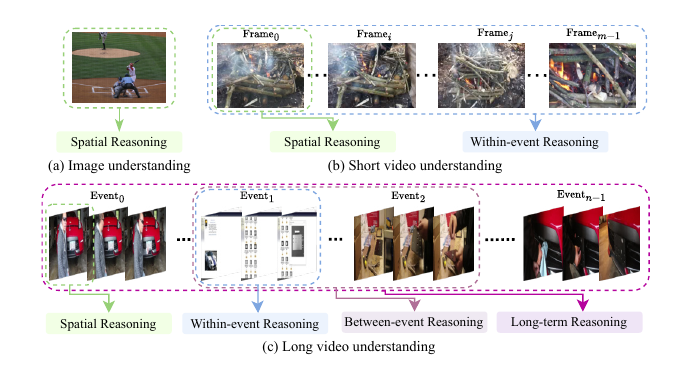}
  \caption{Visual understanding of (a) images, (b) short videos, and (c) long videos.}
  \label{fig:vis_understanding}
\end{figure*}

\section{Long Video Understanding} 
Due to the inherent differences between long video understanding and image or short video understanding, including the presence of various events with more frames and dynamic scenarios, the task of long video understanding presents additional challenges for visual comprehension.

\subsection{Visual Reasoning and Understanding}
\label{sec:2.1}
Visual reasoning demands models to comprehend and interpret visual information and integrate multimodal perception with commonsense understanding \cite{johnson2017clevr, chen2024large}. There are three main types of visual reasoning tasks: visual question answering (VQA), visual captioning (VC) or description (VD), and visual dialog (VDia). VQA \cite{antol2015vqa, zakari2022vqa} involves generating a natural language answer based on the input visual data and accompanying questions. VC and VD systems \cite{vinyals2015show, sharma2018conceptual, li2019visual} typically generate a concise, natural language sentence that summarizes the main content of the visual data and a detailed and comprehensive description of the corresponding visual data, respectively. VDia \cite{das2017visual, qi2020two} involves multi-turn conversations, consisting of a series of question-answer pairs centered around the visual content.

\noindent\textbf{Image understanding.}
As illustrated in Fig. \ref{fig:vis_understanding} (a), image understanding tasks involve a single image for various visual reasoning tasks, such as image captioning and image-centered question answering \cite{sharma2018conceptual, mathew2021docvqa, changpinyo2022all, li2023blip, chen2024spatialvlm}. These tasks focus solely on spatial information, encompassing both coarse-grained understanding \cite{ordonez2011im2text, sohoni2020no} of global visual context and fine-grained understanding \cite{wei2021fine, liu2024llavanext_image, peng2024frih} of local visual details.

\noindent\textbf{Short video understanding.} Unlike image understanding tasks, which involve only static visual data, short video understanding also incorporates temporal information from multiple visual frames \cite{xu2016msr, bain2021frozen, li2023videochat, li2024mvbench}. In addition to spatial reasoning \cite{ranasinghe2024learning}, within-event temporal reasoning and spatiotemporal reasoning across frames play crucial roles for short video understanding \cite{huang2018makes, lin2019tsm, diba2023spatio}.


\noindent\textbf{Long video understanding.} Long videos, spanning minutes or even hours, typically consist of multiple events, encompassing much richer spatial content and temporal variations compared to short videos \cite{mangalam2024egoschema, li2024videovista, song2024moviechat, song2024moviellm}. As summarized in Fig. \ref{fig:vis_understanding} (c), long video understanding involves not only spatial and within-event temporal reasoning but also between-event reasoning and long-term reasoning from different video events \cite{wu2019long, wu2021towards, wang2023selective, zhou2024mlvu, fang2024mmbench}.


\subsection{Challenges of Long Video Understanding}
\label{sec:2.2}
Compared with images and short videos, long-form videos introduce new challenges to comprehensive visual understanding, as follows:

\noindent\textbf{Rich fine-grained spatiotemporal details.} Long videos, which cover a wide range of topics, scenes, and activities, contain varying details such as objects, events, and attributes \cite{fu2024video, wu2024longvideobench}. These details are much richer compared to static images and short videos with multiple similar frames, making long video understanding more challenging. For instance, fine-grained spatial question answering can be introduced in any frame, while temporal question answering can be introduced between or among frames for long video reasoning tasks \cite{song2024moviechat}. MM-LLMs for long video understanding must capture all relevant fine-grained spatiotemporal details from video frames spanning minutes or even hours, using a limited number of visual tokens.

\noindent\textbf{Dynamic events with scene transitions and content changes.} Long videos often contain various dynamic events with significant differences in scenes and content \cite{wu2024longvideobench}. These events can be semantically related and temporally coordinated according to their order of appearance \cite{bao2021dense}, or they can exhibit significant semantic differences due to plot twists \cite{papalampidi2019movie}. Between-event reasoning involving multiple events with diverse visual information is crucial for accurate content understanding \cite{cheng2024enhancing, qian2024momentor}. For MM-LLMs, distinguishing semantic differences and maintaining semantic coherence across varying events are essential for long video understanding.


\noindent\textbf{Long-term correlation and dependencies.} Long videos often contain actions and events that span extended periods. Capturing long-term dependencies and understanding how different parts of the video relate to each other over the long period is challenging \cite{wu2019long}. Video LLMs designed for images or short videos typically fail to contextualize the present event in relation to past or future events that are far from the current time \cite{wu2021towards}, as well as in long-term decision-making \cite{wang2024lvbench}.

\begin{figure*}[t]
  \centering
  \includegraphics[width=0.85\linewidth]{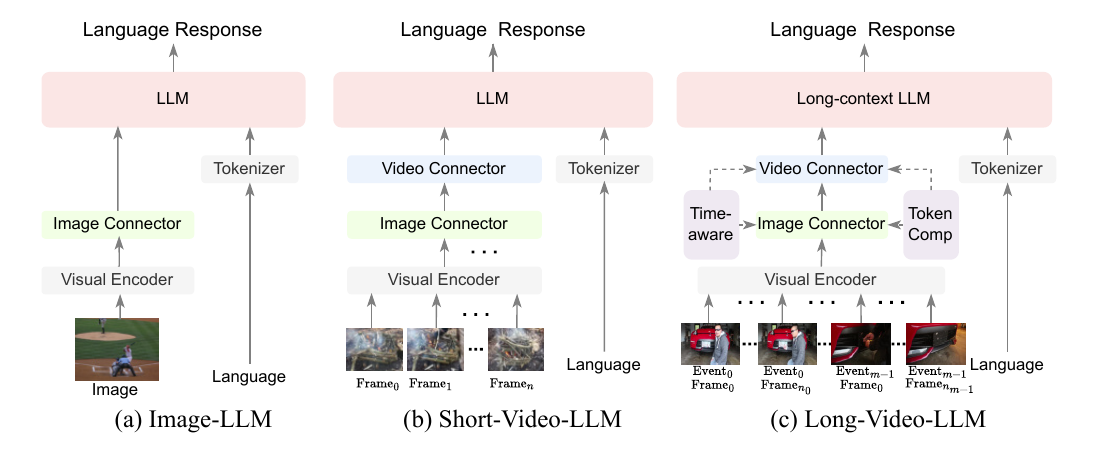}
  \caption{MM-LLMs of (b): Image-LLM, (c) Short-Video-LLM and (c) Long-Video-LLM.}
  \label{fig:model_arc}
\end{figure*}

\begin{table*}[t]
\centering
\resizebox{1.0\linewidth}{!}{
\begin{tabular}{l c | cc c ccc ccc cc}
\toprule
\multirow{2}*{\textbf{Model}} & \multirow{2}*{\textbf{Year}}  & \multicolumn{2}{c}{\textbf{Backbone}} & & \multicolumn{3}{c}{\textbf{Connector}} & \multirow{2}*{\textbf{\#Frame}}  & \multirow{2}*{\textbf{\#Token}} & \multirow{2}*{\textbf{Hardware}} & \multicolumn{2}{c}{\textbf{Training}} \\
\cmidrule{3-4} \cmidrule{6-8} \cmidrule{12-13}
& & Visual Encoder & LLMs & & Image-level & Video-level & Long-video-level & & & & \textbf{PT} & \textbf{IT}\\
\midrule
InstructBLIP \citeyearpar{dai2023instructblip} & 23.05 & EVA-CLIP-ViT-G/14 & FlanT5, Vicuna-7B/13B & & Q-Former & -- & -- & 4 & 32/128 & 16 A100-40G & Y-N-N & Y-N-N \\
VideoChat \citeyearpar{li2023videochat} & 23.05 & EVA-CLIP-ViT-G/14 & StableVicuna-13B &  & Q-Former &  Global multi-head relation aggregator & -- & 8 & /32 & 1 A10 & Y-Y-N & Y-Y-N \\
Video-LLaMA \citeyearpar{zhang2023video} &23.06 & EVA-CLIP-ViT-G/14 & LLaMA, Vicuna & & Q-Former & Q-Former & -- & 8 & /32 & -- & Y-Y-N & Y-Y-N \\
Video-ChatGPT \citeyearpar{maaz2023video} & 23.06 & CLIP-ViT-L/14 & Vicuna1.1-7B & & Spatial-pooling & Temporal-pooling & -- & 100 & /356 & 8 A100-40G & N-N-N & N-Y-N \\
Valley \citeyearpar{luo2023valley} & 23.06 & CLIP-ViT-L/14 & StableVicuna-7B/13B & & -- & Transformer and Mean pooling & -- & 0.5 fps & /256+T & 8 A100 80G & Y-Y-N & Y-Y-N \\
MovieChat \citeyearpar{song2024moviechat}& 23.07 & EVA-CLIP-ViT-G/14 & LLama-7B &  & Q-Former & Frame mergin, Q-Former& Merging adjacent frames & 2048 & 32/32 & -- & E2E & E2E \\
Qwen-VL \citeyearpar{bai2023qwenvl} & 23.08 & Openclip-ViT-bigG & Qwen-7B & & Cross-attention & -- & -- & 4 & /256 & -- & Y-N-N & Y-N-N \\
Chat-UniVi \citeyearpar{jin2024chat} & 23.11 & CLIP-ViT-L/14 & Vicuna1.5-7B & & Token merging & -- & -- & 64  & /112 & -- & Y-N-N & Y-Y-N \\
Video-LLaVA \citeyearpar{lin2023video} & 23.11 & LanguageBind-ViT-L/14 & Vicuna1.5-7B & & -- & -- & -- & 8 & 256/2048 & 4 A100-80G & Y-Y-N & Y-Y-N \\
LLaMA-VID \citeyearpar{li2023llama}& 23.11 & CLIP-ViT-L/14 & Vicuna-7B/13B  & & \multicolumn{3}{c}{Context attention and pooling} & 1 fps & 2/  & 8 A100 & Y-Y-N & Y-Y-Y \\
VTimeLLM \citeyearpar{huang2024vtimellm} & 23.11 & CLIP-ViT-L/14 & Vicuna1.5-7B/13B &  & Frame feature & -- & -- & 100 & 1/100 & 1 RTX-4090 & Y-Y-N  & N-Y-N \\
VideoChat2 \citeyearpar{li2024mvbench} & 23.11 & EVA-CLIP-ViT-G/14 & Vicuna0-7B & & -- &  Q-Former & -- & 16 & /96 & --  & Y-Y-N & Y-Y-N \\
Vista-LLaMA \citeyearpar{ma2023vista} & 23.12 & EVA-CLIP-ViT-G/14 & LLaVa-Vicuna-7B & & Q-Former  & Temporal Q-Former & -- & 16 & 32/512 & 8 A100-80GB & E2E & E2E \\
TimeChat \citeyearpar{ren2024timechat} & 23.12 & EVA-CLIP-ViT-G/14 & LLaMA2-7B & & Q-Former & Sliding window Q-Former & Time-aware encoding & 96 & /96 & 8 V100-32G  & Y-Y-N & N-N-Y \\
VaQuitA \citeyearpar{wang2023vaquita} & 23.12 & CLIP-ViT-L/14 & LLaVA1.5-LLaMA-7B & & -- & Video Perceiver, VQ-Former & -- & 100 & /356 & 8 A100-80GB & E2E & E2E \\
Dolphins \citeyearpar{ma2023dolphins} & 23.12 & CLIP-ViT-L/14 & OpenFlamingo & & \multicolumn{2}{c}{Perceiver Resamplar, Gated cross-attention} & Time embedding & -- & -- & 4 A100 & N-Y-N & Y-Y-N \\
Momentor \citeyearpar{qian2024momentor} & 24.02 & CLIP-ViT-L/14 & LLaMA-7B & & \multicolumn{3}{c}{Frame feature, Temporal Perception Module, Grounded Event-Sequence Modeling} & 300 & 1/300 & 8 A100 & Y-Y-N & N-Y-N \\
MovieLLM \citeyearpar{song2024moviellm} & 24.03 & CLIP-ViT-L/14 & Vicuna-7B/13B  & & \multicolumn{3}{c}{Context attention and pooling} & 1 fps & 2/  & 4 A100 & Y-Y-N & Y-Y-Y \\
MA-LMM \citeyearpar{he2024ma} & 24.04 & EVA-CLIP-ViT-G/14 & Vicuna-7B & & Q-Former & Memory Bank Compression & Merging adjacent frames & 100 & /32 & 4 A100  & E2E & E2E \\
PLLaVA \citeyearpar{xu2024pllava} & 23.04 & CLIP-ViT-L/14 & LLaVA-Next-LLM & & \multicolumn{3}{c}{Adaptive Pooling} & 64 & 2304 & -- & Y-N-N & Y-Y-N \\
LongVLM \citeyearpar{weng2024longvlm} & 23.04 & CLIP-ViT-L/14 & Vicuna1.1-7B & & \multicolumn{3}{c}{Hierarchical token merging} & 100 & /305 & 4 A100 80G & Y-N-N & Y-Y-N \\
MiniGPT4-Video \citeyearpar{ataallah2024minigpt4} & 24.04 & EVA-CLIP-ViT-G/14 & LLaMA2-7B, Mistral-7B & & Merging adjacent tokens & -- & -- & 90 & 64/5760 &  & Y-Y-N  & N-Y-N \\
RED-VILLM \citeyearpar{huang2024image} & 24.04 & Openclip-ViT-bigG & QWen-7B & & Spatial pooling & Temporal pooling & -- & 100 & /1124 &  & Y-N-N  & Y-Y-N \\
ST-LLM \citeyearpar{liu2024st} & 24.04 & BLIP-2 & InstructBLIP-Vicuna1.1-7B  & & Q-Former & Masked video modeling & Global-Local input & 16 & /512 & 8 A100 & E2E & E2E \\
\multirow{2}*{LLaVA-NeXT-Video \citeyearpar{zhang2024llavanextvideo}} & \multirow{2}*{24.04} & \multirow{2}*{CLIP-ViT-L/14} & Vicuna1.5-7B/13B, Mistral-7B &  & \multirow{2}*{Merging adjacent tokens}  & \multirow{2}*{--} \multirow{2}*{--} & \multirow{2}*{--} & \multirow{2}*{32} & \multirow{2}*{4608} & \multirow{2}*{--} & \multirow{2}*{Y-Y-N} & \multirow{2}*{Y-Y-N} \\
&  & & Nous-Hermes-2-Yi-34B \\
Mantis-Idefics2 \citeyearpar{jiang2024mantis} & 24.05 & SigLIP-SO400M & Mistral0.1-7B & & Perceiver resampler & -- & -- & 8 & 64/512 & 16 A100-40G  & Y-N-N & N-Y-N \\
VideoLLaMA 2 \citeyearpar{cheng2024videollama} & 24.06 & CLIP-ViT-L/14 & Mistral-7B-Instruct & & \multicolumn{2}{c}{Spatial-Temporal
Convolution} & -- & 8 & /576 & -- & Y-Y-N & Y-Y-N \\
LongVA \citeyearpar{zhang2024long} & 24.06 & CLIP-ViT-L/14 &  Qwen2-7B-224K & & Merging adjacent tokens & Expanding tokens & -- & 384 & 55,296 & 8× A100-80G & -- & Y-N-N \\
Artemis \citeyearpar{qiu2024artemis} & 24.06 & CLIP-ViT-L/14 &  Vicuna1.5-7B & & \multicolumn{3}{c}{Average pooling} & 5 & /356 & 8 × A800 & Y-Y-N & N-Y-N \\
\multirow{2}*{VideoGPT+ \citeyearpar{maaz2024videogpt+}} & \multirow{2}*{24.06} & CLIP-ViT-L/14 & \multirow{2}*{Phi3-Mini-3.8B} & & \multirow{2}*{Adaptive pooling} & \multirow{2}*{Adaptive pooling} & \multirow{2}*{--} & \multirow{2}*{16} & \multirow{2}*{/2560} & \multirow{2}*{8 × A100 40G}  & \multirow{2}*{Y-Y-N}  & \multirow{2}*{N-Y-N}  \\
&  &  InternVideo-v2  \\
IXC-2.5 \citeyearpar{zhang2024internlm} & 24.07 & CLIP-ViT-L/14-490 & InternLM2-7B  & & Merging adjacent tokens & Expanding tokens & Frame index & 64 & 400/25600 & -- & Y-Y-N & Y-Y-N \\
EVLM \citeyearpar{chen2024evlm} & 24.07 & EVA2-CLIP-E-Plus & Qwen-14B-Chat 1.0 & & Gated cross attention & -- & -- & -- & /16 & -- & Y-Y-N & Y-Y-N \\
SlowFast-LLaVA \citeyearpar{xu2024slowfast} & 24.07 & CLIP-ViT-L/14 & Vicuna1.5-7B & & Merging adjacent tokens & \multicolumn{2}{c}{Slow and fast pathway} & 50 & 3680 & A100-80G & --  & -- \\
LLaVA-NeXT-Interleave \citeyearpar{li2024llava_interleave} & 24.07 & SigLIP-SO400M & Qwen1.5-0.5B/7B/14B & & -- & -- & -- & 16 & 729/11664 & -- & Y-N-N & Y-Y-N \\
Kangaroo \citeyearpar{liu2024kangaroo} & 24.08 & EVA-CLIP-ViT-G/14 & LLaMA3-8B &  & \multicolumn{3}{c}{3D Depthwise convolution} &  & -- & -- &  Y-Y-N & Y-Y-Y \\
VITA \citeyearpar{fu2024vita} & 24.08 & InternViT-300M-448px & Mixtral 8x7B &  & MLP & --  &  & 16 & 256/4096 & -- &  Y-Y-N & Y-Y-N \\
LLaVA-OneVision \citeyearpar{li2024llava} & 24.08 & SigLIP-SO400M & Qwen2-7B & & Merging adjacent tokens & -- & -- & 1fps & 729/ & -- & Y-N-N & Y-Y-N \\
LONGVILA \citeyearpar{xue2024longvila} & 24.08 & -- & -- & & \multicolumn{3}{c}{Multi-Modal Sequence Parallelism} & 1024 & 256/ & 256 A100 80G & Y-Y-N & Y-Y-Y \\
LongLLaVA \citeyearpar{wang2024longllava} & 24.09 & CLIP-ViT-B/32  & LLaVA1.6-13B & & Merging adjacent tokens & Mamba Layers & Hybrid architecture & 256 & 144/ & 24 A800 80G & Y-N-N & Y-Y-N \\
Qwen2-VL \citeyearpar{wang2024qwen2} & 24.09 & CLIP-ViT-L/14 & QWen2-1.5B/7B/72B & & Merging adjacent tokens & 3D convolutions & -- & 2 fps & 66/ & -- & Y-N-N & Y-Y-N \\
\bottomrule
\end{tabular}
}
\caption{Comparison of mainstream Video-LLMs. "PT" and "IT" denote the two stages of pre-training and instruction-tuning during model training. The letters "Y" (Yes) and "N" (No) indicate whether image, short-video, and long-video language datasets are used in these stages. "E2E" stands for an end-to-end training pipeline.
}
\label{tab:models}
\end{table*}

\section{Advances in Model Architecture}
\label{sec:3}
In this section, we discuss the advances of MM-LLMs from image-targeted to long-video-targeted models, from the perspective of model architecture. As illustrated in Fig. \ref{fig:model_arc}, MM-LLMs for images, short videos, and long videos share a similar structure comprising a visual encoder, an LLM backbone, and an intermediary connector. Unlike the image-level connector in image-targeted MM-LLMs, the video-level connector is crucial for integrating cross-frame visual information. In LV-LLMs, designing the connector is more challenging, requiring efficient compression of amounts of visual information and incorporating temporal knowledge to manage long-term correlations.

\subsection{Visual Encoder and LLM Backbone}
MM-LLMs, encompassing both image-targeted and video-targeted models, typically utilize similar visual encoders for visual information extraction. LLM backbones are also universal in early MM-LLM methods, while existing LV-LLMs tend to use long-context LLMs in the implementation.

\noindent\textbf{Visual encoder.} The pretrained visual encoders are responsible for capturing vision knowledge from raw visual data. As summarized in Table \ref{tab:models}, image encoders like CLIP-ViT-L/14 \cite{radford2021learning}, EVA-CLIP-ViT-G/14 \cite{sun2023eva}, OpenCLIP-ViT-bigG/14 \cite{cherti2023reproducible}, and SigLIP-SO400M \cite{zhai2023sigmoid} are widely utilized as visual modality encoders in image- and video-targeted LLMs. Recent work \cite{li2024llavanext-ablations} shows that the visual representation, including image resolution, the size of visual token, and the pre-training visual resources, play a more important role than the size of the visual encoder.

\noindent\textbf{LLM backbone.} The LLM is the core module in visual understanding systems, inheriting properties of reasoning and decision-making. Compared to closed-source LLMs like GPT-3/4 \cite{brown2020language, achiam2023gpt} and Gemini-1.5 \cite{reid2024gemini}, various open-source LLMs are more commonly used in implementing visual LLMs. These include Flan-T5 \cite{chung2024scaling}, LLaMA \cite{touvron2023llama1, touvron2023llama2, dubey2024llama}, Vicuna \cite{vicuna2023}, QWen \cite{bai2023qwen}, Mistral \cite{jiang2023mistral}, Openflamingo \cite{awadalla2023openflamingo}, Yi \cite{young2024yi}, and InternLM \cite{team2023internlm, cai2024internlm2}. 

\noindent The strength of the LLM typically correlates with superior multimodal capabilities in visual LLMs \cite{li2024llavanext-strong, li2024llavanext-ablations}. This means that, for LLMs of the same scale, those with better language capabilities perform better, and for the same LLMs with different model sizes, larger models tend to yield better multimodal performance. Additionally, long-context LLMs that extend the context length to hundreds of thousands of tokens support learning with more extensive data \cite{yang2024qwen2}. Recent LV-LLMs effectively transfer the LLM's long-context understanding ability to the vision modality \cite{zhang2024long}.

\subsection{Modality Interface}
The connectors between visual encoders and LLMs serve as a modality interface, mapping visual features to the language feature space. Given the variability in visual data sources, these connectors can be categorized into image-level, video-level, and long-video-level connectors.

\noindent\textbf{Image-level connectors.} 
Image-level connectors are used to map image features to the language space for processing raw visual tokens, and they are widely used in both image-targeted and video-targeted MM-LLMs. These connectors can be categorized into three groups: \textbf{The first group} directly uses a single linear layer \cite{liu2024visual} or a multi-layer perceptron (MLP) \cite{liu2024improved} to map image features into the language embedding space. However, this method, which retains all visual tokens, is not suitable for visual understanding tasks involving multiple images. To address the limitations of retaining all visual tokens, \textbf{the second group} employs various pooling-based methods. These include spatial pooling \cite{maaz2023video}, adaptive pooling \cite{xu2024pllava}, semantic-similar token merging \cite{jin2024chat}, and adjacent token averaging \cite{zhang2024llavanextvideo, li2024llava}. \textbf{The third group} utilizes cross-attention or transformer-based structures, such as Q-Former \cite{li2023blip} and Perceiver Resampler \cite{jaegle2021perceiver}, for image feature compression. Q-Former is a lightweight transformer structure that employs a set of learnable query vectors to extract and compress visual features. Many visual LLMs \cite{dai2023instructblip, li2023videochat, ma2023vista, liu2024st}, following BLIP-2, choose the Q-Former-based connector. Other visual LLMs \cite{ma2023dolphins, jiang2024mantis} opt for the Perceiver Resampler to reduce computational burden by extracting patch features.

\noindent\textbf{Video-level connectors.} 
Video-level connectors are used for extracting sequential visual data and further compressing visual features. Compared to the solely image-level connectors in image-targeted MM-LLMs, video-level connectors are essential for video-targeted MM-LLMs, including LV-LLMs. Some methods directly concatenate image tokens before inputting them to the LLMs, making them sensitive to the number of frame images \cite{dai2023instructblip, lin2023video}. Similar structures used for token compression in image-level connectors can be adapted for video-level interfaces, such as pooling-based and transformer-based structures. Pooling along the time series dimension is a straightforward way to reduce temporal information redundancy \cite{maaz2023video, song2024moviechat}. Transformer-based methods, such as Video Q-Former \cite{zhang2023video, ma2023vista, ren2024timechat} and Video Perceiver \cite{wang2023vaquita}, are effective in extracting video features while reducing data complexity. Additionally, 3D-Convolution-based methods can extract and compress visual data from both the spatial and temporal dimensions \cite{cheng2024videollama, liu2024kangaroo}.

\noindent\textbf{Long-video-level connectors.} The connectors specifically designed for long-video LLMs incorporate two special considerations: efficient visual information compression for handling long-form visual data and time-aware design for preserving temporal information.

\noindent Efficiently compressing visual information requires not only reducing the input visual tokens to an acceptable quantity but also preserving the complete spatiotemporal details contained in long videos. Videos contain two types of data redundancy: spatial data redundancy within frames and spatiotemporal data redundancy across frames \cite{li2022hybrid, chen2023survey}. On the one hand, spatial data redundancy arises when region-level pixels within frames are the same, leading to inefficiencies when representing the redundant visual frame through full visual tokens. To reduce spatial video data redundancy, the LLaVA-Next-series methods \cite{zhang2024llavanextvideo, li2024llava_interleave, liu2024llavanext_image, li2024llava} merge adjacent frame patch tokens, and Chat-UniVi \cite{jin2024chat} merges similar frame patch tokens. On the other hand, spatiotemporal data redundancy includes both cross-frame pixel redundancy and motion redundancy \cite{pourreza2023boosting}, where the semantic information is similar among these redundant video frames. To reduce spatiotemporal video redundancy, MovieChat \cite{song2024moviechat} and MA-LMM \cite{he2024ma} merge frame features with higher frame similarity before inputting them to LLMs. In addition to reducing redundant information, preserving more video spatiotemporal details is crucial for accurate long video reasoning \cite{diba2023spatio}. To balance global and local visual information and support more frame inputs, SlowFast-LLaVA \cite{xu2024slowfast} employs a slow pathway to extract features at a low frame rate while retaining more visual tokens, and a fast pathway at a high frame rate with a larger spatial pooling stride to focus on motion cues.



\noindent Additionally, time-involved visual data efficiently manage the temporal and spatial information inherent in long-form videos \cite{hou2024memotichat}. The time-aware design can enhance the temporal-capturing capability of video-related LLMs, which is particularly beneficial for long video understanding. Both VTimeLLM \cite{huang2024vtimellm} and InternLM-XComposer-2.5 (IXC-2.5) \cite{zhang2024internlm} use frame indices to enhance temporal relations. The difference lies in their approach: VTimeLLM learns temporal information by training with decoded text that includes frame indices, while IXC-2.5 encodes frame indices along with the frame image context. TimeChat \cite{ren2024timechat} and Momentor \cite{qian2024momentor} inject temporal information directly into frame features for fine-grained temporal information capture. Specifically, TimeChat designs a Time-aware Frame Encoder to extract visual features with corresponding timestamp descriptions at the frame level, while Momentor utilizes a Temporal Perception Module for continuous time encoding and decoding, injecting temporal information into frame features.

\section{Advances in Model Training}
\label{sec:4}
Multimodal LLMs for visual understanding consist of two principal stages: pre-training (PT) for vision-language feature alignment and instruction-tuning (IT) for instruction-aware response. 

\begin{figure*}[t]
  \centering
  \includegraphics[width=0.85\linewidth]{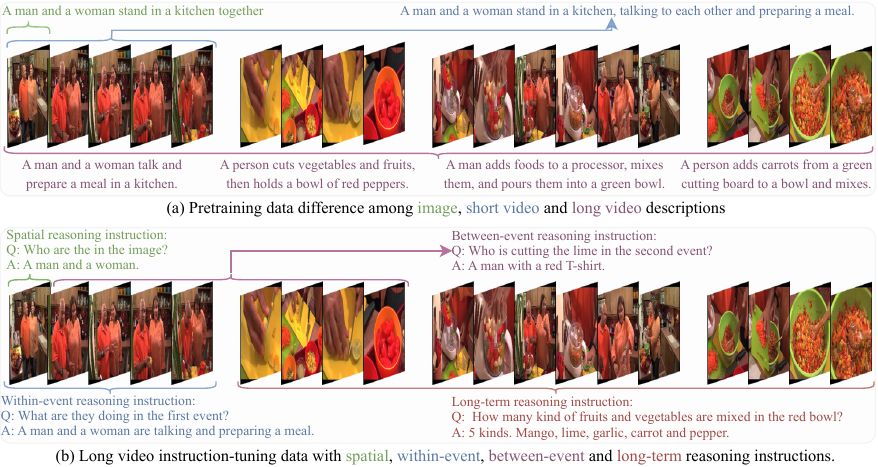}
  \caption{Long video sample for pretraining and instruction-tuning.}
  \label{fig:long_video_dataset}
\end{figure*}

\subsection{Pre-training}
Vision-language pre-training for MM-LLMs aims to align visual features with the language space using text-paired data. This includes pre-training with image-, short-video-, and long-video-text datasets.

\noindent Initially introduced for visual LLMs focused on images, \textbf{image-text pre-training} is also widely used in video-related understanding tasks. Coarse-grained image-text pair datasets, such as COCO Captions \cite{chen2015microsoft} and CC-3M \cite{sharma2018conceptual}, are employed for global vision-language alignment. Fine-grained image-text datasets like ShareGPT4V-PT \cite{chen2023sharegpt4v} are used for locally spatial semantics alignment. Given the limited changes in semantic content of short videos, short-video-text paired datasets, such as Webvid-2M \cite{bain2021frozen}, can be used similarly for \textbf{short-video-text pre-training}. Similarly, \textbf{long-video-text pre-training} is important to capture the temporal semantic alignment of long videos for long video understanding. Given the absence of long-term cross-modal correlation in image-text and short-video-text pairs, long-video-text pre-training datasets with pairs of long videos and their corresponding text descriptions are necessary \cite{argaw2023long}. Moreover, as shown in Fig. \ref{fig:long_video_dataset} (a), the scenes and events in long videos vary significantly across frames, necessitating event-level vision-language alignment \cite{qian2024momentor} for long-video-text pre-training, which is markedly different from both image-text and short-video-text pre-training \cite{zhang2024long}.

\subsection{Instruction-tuning}
Instruction-tuning with vision-language sources enables LLMs to follow instructions and generate human-like text. Multimodal vision-language instruction-following data \cite{dai2023instructblip, liu2024visual}, including both image-text and video-text pairs, are used to align multimodal LLMs with human intent, thereby enhancing their ability to complete real-world tasks.

\noindent Similar to the pre-training stage, \textbf{image-text instruction-tuning} is also employed in various vision-understanding tasks, including image, short-video, and long-video understanding tasks. Basic image-based instruction-following datasets, such as ShareGPT4V-Instruct \cite{chen2023sharegpt4v} and LLaVA-Instruct \cite{liu2024visual}, provide high-quality instruction-tuning data for basic spatial reasoning and chat capabilities. For video-related LLMs,  \textbf{short-video-text instruction-tuning} is necessary to enable multimodal LLMs to understand temporal sequences, as seen in models like Video-ChatGPT \cite{maaz2023video} and VideoChat \cite{li2023videochat}. Short-video-LLMs require both spatial and within-event reasoning instructions to understand the spatial and small-scale temporal content of short videos. However, the limited content and semantic changes in short videos are insufficient for long video understanding tasks, where frames are more numerous and exhibit significant variation.  \textbf{Long-video-text instruction-tuning} is specifically introduced to better capture and understand long videos. In addition to spatial and within-event reasoning instructions, between-event and long-term reasoning instructions are necessary for the comprehensive understanding of long videos, as shown in Fig. \ref{fig:long_video_dataset} (b). Among the introduced long-video instruction-format datasets, Long-VideoQA \cite{li2023llama} and Video-ChatGPT \cite{maaz2023video} are not time-aware, containing only long videos with corresponding data. VTimeLLM \cite{huang2024vtimellm}, TimeIT \cite{ren2024timechat}, and Moment-10M \cite{qian2024momentor} are time-aware, incorporating extra temporal information to enhance temporal correlation.


\begin{table*}[t]
\centering
\resizebox{1.0\linewidth}{!}{
\begin{tabular}{l cc | ccc ccccc ccc}
\toprule
\multirow{2}*{\textbf{Model}} & \multirow{2}*{\textbf{LLM}} & \multirow{2}*{\textbf{Long}} & \textbf{TGIF-QA} & \textbf{MSVD-QA}& \textbf{MSRVTT-QA} & \textbf{NeXT-QA} & \textbf{ActivityNet-QA} & \multicolumn{6}{c}{\textbf{GPT-based Evaluation}(~2mins)}  \\
\cmidrule{9-14}
& & & (2-5s) & (10-15s) & (10-15s) & (42.9s) & (~2mins) & \textbf{CI} & \textbf{DO} & \textbf{CU} & \textbf{TU} & \textbf{CO} & \textbf{Average}\\
\midrule
InstructBLIP & Vicuna-7B & \textbf{\xmark} & -- & 41.8/ & 22.1/ & -- & -- & -- & -- & -- & -- & -- & -- \\
Video-ChatGPT & Vicuna1.1-7B & \textbf{\xmark} & 51.4/3.0 & 64.9/3.3 & 49.3/2.8 & -- & 35.2/2.8 & 2.40 & 2.52 & 2.62 & 1.98 & 2.37 & 2.38 \\
MA-LMM & Vicuna-7B & \textbf{\cmark} & -- & 60.6/ & 48.5/- & -- & 49.8/ & -- & -- & -- & -- & -- & -- \\
Valley & StableVicuna-7B & \textbf{\xmark} & -- & 60.5/3.3 & 51.1/2.9 & -- & 45.1/3.2 & 2.43 & 2.13 & 2.86 & 2.04 & 2.45 & 2.38 \\
MovieLLM & Vicuna-7B & \textbf{\cmark} & -- & 63.2/3.5 & 52.1/3.1 & -- & 43.3/3.3 & 2.64 & 2.61 & 2.92 & 2.03 & 2.43 & 2.53 \\
Vista-LLaMA & Vicuna-7B & \textbf{\xmark} & -- & 65.3/3.6 & 60.5/3.3 & 60.7/3.4 & 48.3/3.3 & 2.44  & 2.31 & 2.64 & 3.18 & 2.26 & 2.57 \\
RED-VILLM & LLaVA-7B & \textbf{\xmark} & 55.9/3.1 & 68.9/2.8 & 52.4/2.9 & -- & 39.2/3.0 & 2.57 & 2.64 & 3.13 & 2.21 & 2.39 & 2.59 \\
Momentor & LLaMA-7B & \textbf{\xmark} & -- & 68.9/3.6 & 55.6/3.0 & -- & 40.8/3.2 & -- & -- & -- & -- & -- & -- \\
Video-LLaVA & Vicuna1.5-7B & \textbf{\xmark} & 70.0/4.0 & 70.7/3.9 & 59.2/3.5 & -- & 45.3/3.3 & -- & -- & -- & -- & -- & -- \\
Artemis & Vicuna1.5-7B  & \textbf{\cmark} & -- & 72.1/3.9 & 56.7/3.2 & -- & 39.3/2.9 & 2.69 & 2.55 & 3.04 & 2.24 & 2.70 & 2.64 \\
MovieChat & LLaMA-7B & \textbf{\cmark} & -- & 75.2/3.8 & 52.7/2.6 & -- & 45.7/3.4 & 2.76 & 2.93 &  3.01 & 2.24 & 2.42 & 2.67 \\
VaQuitA & LLaMA-7B & \textbf{\xmark} & -- & 74.6/3.7 & 68.6/3.3 & -- & 48.8/3.3 & -- & -- & -- & -- & -- & -- \\
RED-VILLM & QWen-VL-7B & \textbf{\xmark} & 62.3/3.3 & 71.2/3.7 & 53.9/3.1 & -- & 44.2/3.2 & 2.69 & 2.72 & 3.32 & 2.32 & 2.47 & 2.70 \\
MiniGPT4-Video & Mistral-7B & \textbf{\xmark} & 72.2/4.1 & 73.9/4.1 & 58.3/3.5 & -- & 44.3/3.4 & 2.97 & 2.58 & 3.17 & 2.38 & 2.44 & 2.71 \\
VTimeLLM & Vicuna-7B & \textbf{\xmark} & -- & -- & -- & -- & -- & 2.49 & 2.78 & 3.10 & 3.40 & 2.47 & 2.85 \\
MiniGPT4-Video & LLaMA2-7B & \textbf{\xmark} & 67.9/3.7 & 72.9/3.8 & 58.8/3.3 & -- & 45.9/3.2 & 2.93 & 2.97 & 3.45 & 2.47 & 2.60 & 2.88 \\
Chat-UniVi & Vicuna1.5-7B & \textbf{\xmark} & 69.0/3.8 & 69.3/3.7 & 55.0/3.1 & -- & 46.1/3.3 & 2.89 & 2.91 & 3.46 & 2.40 & 2.81 & 2.89 \\
LLaMA-VID & Vicuna-7B & \textbf{\cmark} & -- & 69.7/3.7 & 57.7/3.2 & -- & 47.4/3.3 & 2.96 & 3.00 & 3.53 & 2.46 & 2.51 & 2.89 \\
LongVLM & Vicuna1.1-7B & \textbf{\cmark} & -- & 70.0/3.8 & 59.8/3.3 & -- & 47.6/3.3 & 2.76 & 2.86 & 3.34 & 2.39  & 3.11 & 2.89 \\
VideoChat2 & Vicuna0-7B & \textbf{\xmark} & -- & 70.0/3.9 & 54.1/3.3 & -- & 49.1/3.3 & 3.02 & 2.88 & 3.51 & 2.66 & 2.81 & 2.98 \\
SlowFast-LLaVA & Vicuna1.5-7B & \textbf{\cmark} & 78.7/4.2 & 79.1/4.1 & 65.8/3.6 & 64.2/ & 56.3/3.4 & 3.09 & 2.70 & 3.57 & 2.52 & 3.35 & 3.04 \\
PLLaVA & LLaVA-Next-7B & \textbf{\cmark} & 77.5/4.1 & 76.6/4.1 & 62.0/3.5 & -- & 56.3/3.5 & 3.21 & 2.86 & 3.62 & 2.33  & 2.93 & 3.12 \\
VideoLLaMA2-16 & Mistral-7B-Instruct & \textbf{\xmark} & -- & 70.9/3.8 & -- & -- & 50.2/3.3 & 3.16 & 3.08 & 3.69 & 2.56 & 3.14 & 3.13 \\
VideoLLaMA2-8 & Mistral-7B-Instruct & \textbf{\xmark} & -- & 71.7/3.9 & -- & -- & 49.9/3.3 & 3.09 & 3.09 & 3.68 & 2.63 & 3.25 & 3.15 \\
ST-LLM & Vicuna-7B & \textbf{\cmark} & -- & 74.6 /3.9 & 63.2/3.4 & -- & 50.9/3.3 & 3.23  & 3.05 & 3.74 & 2.93 & 2.81 & 3.15 \\
LongVA-32 & Qwen2-7B-224K & \textbf{\cmark} & -- & -- & -- & 67.1/ & /2.8  & 3.65 & 3.08 & 3.10 & 3.74 & 2.28 & 3.17 \\
LongVA-64 & Qwen2-7B-224K & \textbf{\cmark} & -- & -- & -- & 68.3/ & /2.8  & 3.64 & 3.05 & 3.09 & 3.77 & 2.44 & 3.20 \\
LLaVA-NeXT-Video & Vicuna1.5-7B & \textbf{\xmark} & -- & -- & -- & -- & 53.5/3.2  & 3.39 & 3.29 & 3.92 & 2.60 & 3.12 & 3.26 \\
LLaVA-NeXT-Interleave & Qwen1.5-7B & \textbf{\xmark} & -- & -- & -- & 78.2 & 55.3/3.13  & 3.51 & 3.28 & 3.89 & 2.77 & 3.68 & 3.43 \\
LLaVA-OneVision & Qwen2-7B & \textbf{\xmark} & -- & -- & -- & -- & 56.6/ & -- & -- & -- & -- & -- & 3.49 \\
LongVA-32-DPO & Qwen2-7B-224K & \textbf{\cmark} & -- & -- & -- & 69.3/ & /2.8  & 4.07 & 3.55 & 3.32 & 4.09 & 2.86 & 3.58 \\
LLaVA-NeXT-Video-DPO & Vicuna1.5-7B & \textbf{\xmark} & -- & -- & -- & -- & 60.2/3.5 & 3.64 & 3.45 & 4.17 & 2.95 & 4.08 & 3.66 \\
\midrule
InstructBLIP & Vicuna-13B & \textbf{\xmark} & -- & 41.2/ & 24.8/ & -- & -- & -- & -- & -- & -- & -- & -- \\
LLaMA-VID & Vicuna-13B & \textbf{\cmark} & -- & 70.0/3.7 & 58.9/3.3 & -- & 47.5/3.3 & 3.07 & 3.05 & 3.60 & 2.58 & 2.63 & 2.99 \\
PLLaVA & LLaVA-Next-13B & \textbf{\cmark} & 77.8/4.2 & 75.7/4.1 & 63.2/3.6 & -- & 56.3/3.6 & 3.27 & 2.99 & 3.66 & 2.47 & 3.09 & 3.27 \\ 
LLaVA-NeXT-Interleave & Qwen1.5-14B & \textbf{\xmark} & -- & -- & -- & 79.1  & 56.2/3.19  & 3.65 & 3.37 & 3.98 & 2.74 & 3.67 & 3.48 \\
LLaVA-NeXT-Interleave-DPO & Qwen1.5-14B & \textbf{\xmark} & -- & -- & -- & 77.9 & 55.0/3.13  & 3.99 & 3.61 & 4.24 & 3.19 & 4.12 & 3.83 \\
\midrule
SlowFast-LLaVA & Nous-Hermes-2-Yi-34B & \textbf{\cmark} & 80.6/4.3 & 79.9/4.1 & 67.4/3.7 & -- & 59.2/3.5 & 3.48 & 2.96 & 3.84 & 2.77 & 3.57 & 3.32 \\
LLaVA-NeXT-Video & Nous-Hermes-2-Yi-34B & \textbf{\xmark} & -- & -- & -- & -- & 58.8/3.4  & 3.48 & 3.37 & 3.95 & 2.64 & 3.28 & 3.34 \\
PLLaVA & LLaVA-Next-34B & \textbf{\cmark} & 80.6/4.3 & 79.9/4.2 & 68.7/3.8 & -- & 60.9/3.7 & 3.60 & 3.20 & 3.90 & 2.67 & 3.25 & 3.48 \\
LLaVA-NeXT-Video-DPO & Nous-Hermes-2-Yi-34B & \textbf{\xmark} & -- & -- & -- & -- & 64.4/3.6 & 3.81 & 3.55 & 4.24 & 3.14 & 4.12 & 3.77 \\
\midrule
LLaVA-OneVision & Qwen2-72B & \textbf{\xmark} & -- & -- & -- & -- & 62.3/ & -- & -- & -- & -- & -- & 3.62 \\
\bottomrule
\end{tabular}
}
\caption{Comparison of mainstream Video-LLMs on video understanding benchmarks of different lengths. Methods with \textbf{\cmark} in the "Long" column are designed for long videos.
}
\label{tab:video_benchmarks}
\end{table*}

\begin{table*}[t]
\centering
\resizebox{1.0\linewidth}{!}{
\begin{tabular}{l cc | ccc ccccc}
\toprule
\multirow{2}*{\textbf{Model}} & \multirow{2}*{\textbf{LLM}} & \multirow{2}*{\textbf{Long}} & \textbf{VideoVista} & \textbf{QVHighlights}   & \textbf{MMBench-Video} & \textbf{EgoSchema} & \textbf{LongVideoBench} & \textbf{MLVU} & \textbf{Video-MME} &\textbf{LVBench} \\
& & & (131s) & (150s) & (165s) & (180s) & (473s) & (12mins) & (1024s) & (4101s)\\
\midrule
Momentor & LLaMA-7B & \textbf{\xmark} & -- & 7.6/17.0 & -- & -- & -- & -- & -- &  \\
TimeChat & LLaMA2-7B & \textbf{\cmark} & -- & 14.5/23.9 & -- & 33.0$^{\dagger}$ & -- & 30.9$^{\dagger}$  & -- & 22.3$^{\spadesuit}$ \\
LLaMA-VID & Vicuna-7B & \textbf{\cmark} & 56.87$^{\ddagger}$ & -- &  -- & 38.5$^{\dagger}$ & -- & 33.2$^{\dagger}$ & -- & 23.9$^{\spadesuit}$ \\
LLaVA-NeXT-Video & Vicuna1.5-7B & \textbf{\xmark} &56.66$^{\ddagger}$ & -- & -- & 43.9$^{\dagger}$ & 43.5$^{\heartsuit}$ & -- & -- \\
VideoLLaMA 2 (16) & Mistral-7B-Instruct & \textbf{\xmark} & 60.47$^{\ddagger}$ & -- & -- & 51.7 & -- & 48.5$^{\dagger}$ & 47.9/50.3$^{\clubsuit}$ & -- \\
PLLaVA & LLaVA-Next-7B & \textbf{\cmark} & 60.36$^{\ddagger}$ & -- & 1.03$^{\dagger}$ & 54.4$^{\dagger}$ & 39.2$^{\heartsuit}$ & -- \\
LongVA & Qwen2-7B-224K & \textbf{\cmark} & 67.36$^{\ddagger}$ & -- & -- & -- & -- & 56.3$^{\dagger}$ & 52.6/54.3$^{\clubsuit}$ & --  \\
IXC-2.5-7B & InternLM2-7B & \textbf{\xmark} & 68.91$^{\ddagger}$ & -- & 1.41 & -- & -- & 58.8 & 55.8/- \\
Kangaroo & LLaMA3-8B & \textbf{\cmark} & 69.50$^{\ddagger}$  & -- & 1.44 & 62.7 & 54.8 & 61.0  & 56.0/57.6$^{\clubsuit}$ & 39.4 \\
\bottomrule
\end{tabular}
}
\caption{Comparison of mainstream Long-Video-LLMs on long video understanding benchmarks. Results with $^{\ddagger}$ are from the VideoVista benchmark \cite{li2024videovista}. Results with $^{\dagger}$ are from the Kangaroo \cite{liu2024kangaroo}. Results with $^{\clubsuit}$ are from Video-MME benchmark \cite{fu2024video}. Results with $^{\spadesuit}$  are from LVBench \cite{wang2024lvbench}. Results with $^{\heartsuit}$ are from LongVideoBench \cite{wu2024longvideobench}.
}
\label{tab:long_benchmarks}
\end{table*}

\section{Evaluation, Performance and Analysis}
\label{sec:5}
In this section, we present a performance comparison across popular evaluation datasets featuring videos of varying lengths, along with our analysis. The comparison is conducted from two perspectives: First, we evaluate video understanding methods on tasks with video lengths ranging from seconds to minutes. Second, we specifically compare performance on extra-long video datasets, with video lengths ranging from minutes to hours.

\subsection{Video Understanding: Seconds to Minutes}
\label{sec:5.1}
As shown in Table \ref{tab:video_benchmarks}, we summarize the general video understanding performance of various visual LLMs on open-ended video question answering benchmarks, including TGIF-QA \cite{jang2017tgif}, MSVD-QA, MSRVTT-QA \cite{xu2017video}, NEXT-QA \cite{xiao2021next}, and ActivityNet-QA \cite{yu2019activitynet}. Additionally, we consider the VideoChatGPT-introduced video-based generative performance benchmark \cite{maaz2023video}, which evaluates five aspects of video-based text generation: Correctness of Information (CI), Detail Orientation (DO), Context Understanding (CU), Temporal Understanding (TU), and Consistency (CO). The video benchmarks with lengths shorter than 1 minute, such as TGIF-QA, MSVD-QA, MSRVTT-QA, and NEXT-QA, are commonly used for short video understanding. In contrast, benchmarks exceeding one minute, such as ActivityNet-QA and the ActivityNet-200-based \cite{caba2015activitynet} generative performance benchmark, are used for long video understanding.

\noindent By comparing the performance in Table \ref{tab:video_benchmarks}, we can conclude that long video understanding is challenging, with the following findings: \textbf{(1)} Video reasoning with more frames introduces more complex visual information and is more challenging. Methods designed to support long videos, such as LongVA \cite{zhang2024long}, show better performance compared to being fed with fewer frames on the same video dataset. However, performance decreases when being fed with more frames from the same video dataset for methods without special designs for long videos, like VideoLLaMA2 \cite{cheng2024videollama}. \textbf{(2)} Short video understanding methods that perform well on seconds-level video understanding often do not perform well on minutes-level moderately long video understanding, such as RED-VILLM \cite{huang2024image} and MiniGPT4-Video \cite{ataallah2024minigpt4}. Long video understanding methods tend to share consistently good performance on both short and moderately long video benchmarks, such as ST-LLM \cite{liu2024st}, SlowFast-LLaVA \cite{xu2024slowfast}, PLLaVA \cite{xu2024pllava}, and MovieChat \cite{song2024moviechat}. This improvement likely stems from better-captured spatiotemporal information in specially designed long video understanding methods. 

\subsection{Video Understanding: Minutes to Hours}
\label{sec:5.2}
To address the unique characteristics of long videos, several long video benchmarks have been introduced in recent years, with video lengths varying from hundreds of seconds to thousands of seconds. EgoSchema \cite{mangalam2024egoschema} and QVHighlights \cite{lei2021detecting} are long-form video understanding datasets designed for multiple-choice question answering and highlight detection, respectively, after accessing all frames. VideoVista \cite{li2024videovista}, MMBench-Video \cite{fang2024mmbench}, and MLVU \cite{zhou2024mlvu} cover various topics and are designed for fine-grained capability evaluation. LongVideoBench \cite{wu2024longvideobench} introduces referring reasoning questions to address the longstanding issue of single-frame bias in long videos. Video-MME \cite{fu2024video} and LVBench \cite{wang2024lvbench} contain numerous hour-level videos. Video-MME further categorizes them into short, medium, and long categories, while LVBench aims to challenge models to demonstrate long-term memory and extended comprehension capabilities.

As shown in Table \ref{tab:long_benchmarks}, we further compare and analyze the performance of long video understanding methods, specifically summarizing their performance on long video benchmarks with lengths varying from hundreds of seconds to thousands of seconds. Unlike the findings in Sec. \ref{sec:5.1}, long video understanding methods typically outperform short video understanding methods. This indicates that specially designed, powerful video-level connectors are essential for long video understanding.

Additionally, the performance on benchmarks with longer video lengths is generally worse than on those with shorter lengths. For example, the performance of methods across VideoVista and MLVU, Video-MME and LVBench, using the same evaluation metric, shows a decline as video length increases. This suggests that long video understanding remains a challenging research topic.

\section{Future Directions}
\label{sec:6}
As summarized above, existing long video understanding methods are less effective than image or short video understanding methods. To meet the demands of an AI-driven society with increasingly more and longer multimodal data, developing more powerful visual LLMs for long video understanding is crucial. The following considerations should be taken into account.

\subsection{More Long Video Training Resources}
The two-stage training pipeline—comprising cross-modal alignment pre-training and visual-language-format instruction tuning—is widely employed for training MM-LLMs \cite{dai2023instructblip, liu2024visual}. However, fine-grained long-video-language training pairs are lacking compared to the commonly used image-language and short-video-language pairs during cross-modal alignment learning \cite{song2024moviellm, qiu2024artemis}. Methods that rely on image-language and short-video-language resources cannot capture long-term correlations during the pre-training stage \cite{zhang2024long}. Additionally, the video length of newly introduced long-video instruction data is limited to the minute level, which significantly restricts the application scenarios for effective reasoning from long video understanding \cite{li2023llama}. Therefore, both long-video-language paired pre-training datasets and long-video-instruction datasets, featuring much longer videos at the hour level with high-quality annotations, need to be created.


\subsection{More Challenging Long Video Understanding Benchmarks}
Various video understanding benchmarks were summarized in the previous section, most of which have been introduced recently. However, these benchmarks focus on one or several aspects of long video understanding, such as LongVideoBench for long-context interleaved video understanding \cite{wu2024longvideobench}, QVHighlights for language-based video highlight understanding \cite{lei2021detecting}, and VideoVista \cite{li2024videovista} and MLVU \cite{zhou2024mlvu} for fine-grained video understanding. Comprehensive long video benchmarks that cover frame-level and segment-level reasoning with time and language are necessary but are currently unexplored for a thorough evaluation of general long video understanding methods \cite{wu2024longvideobench}. Additionally, existing benchmarks are usually at the minute level and cannot adequately test methods' long-term capabilities. Long video understanding methods often suffer from catastrophic forgetting and loss of spatiotemporal details when reasoning with extensive sequential visual information \cite{wang2024lvbench}, such as from hour-level videos. Finally, most existing long video understanding benchmarks focus solely on the visual modality. Incorporating multimodal data, including additional audio and language, would undoubtedly benefit long video understanding tasks.

\subsection{Powerful and Efficient Frameworks} Visual LLMs for videos need to support more visual frames and preserve more visual details with a fixed number of visual tokens. There are four main considerations for the implementation of LV-LLMs: \textbf{(1)} Select long-context LLMs as the LLM backbones. Previous methods have suffered from the context capacity of LLMs and have had to specifically fine-tune an LLM to support more tokens \cite{zhang2024long}. Recent long-context LLMs, such as QWen2 \cite{yang2024qwen2}, LLaMA-3.1 \cite{dubey2024llama}, and DeepSeek-V2 \cite{deepseekv2}, share a context window length of 128K and could be utilized in the design of LV-LLMs. \textbf{(2)} Compress visual tokens more efficiently and with less information loss. Some existing methods face the problem of insufficient compression, such as Chat-UniVi \cite{jin2024chat}, which uses multi-scale token merging, and LongVA \cite{zhang2024long}, which merges adjacent tokens only. Other methods compress too much visual information, such as LLaMA-VID \cite{li2023llama}, which uses context and content tokens, and MA-LMM \cite{he2024ma}, which merges similar frame tokens, resulting in a significant loss of frame details. New frameworks designed for long videos must efficiently compress visual tokens to support more temporal frames and preserve more spatiotemporal details in comprehensive long video understanding tasks. \textbf{(3)} Incorporate extra time-aware designs \cite{ren2024timechat, qian2024momentor} to enhance video reasoning by incorporating temporal information, thereby improving temporal information extraction in long video understanding performance.  \textbf{(4)} Utilize infrastructure capable of supporting memory-intensive long-context training \cite{xue2024longvila}, providing the ability to feed more visual data when equipped with a large number of GPU devices.

\subsection{More Application Scenarios} 
Long video understanding with large models faces several key challenges for more long video applications. Contextual understanding is critical, as long videos require models to maintain temporal coherence and contextual awareness over extended periods \cite{he2024ma}. Real-time processing \cite{karim2024real} is essential for applications like surveillance, live event analysis, and embodied AI, necessitating the development of low-latency models capable of processing video streams in real-time. Multi-modal integration is another frontier, as long videos often contain audio, text, and visual information \cite{zhang2023video, cheng2024videollama}. Future models should better integrate these modalities to enhance understanding and provide a more holistic analysis of video content. 


\section{Conclusion}
In this paper, we summarize the advances of visual LLMs from images to long videos. Based on the analysis of the task differences among image understanding, short video understanding, and long video understanding, we identify the key challenges of long video learning. These challenges include capturing more fine-grained spatiotemporal details and long-term dependencies within compressed visual information from dynamic sequential events with scene transitions and content changes. We then introduce the advances in model architecture and model training from Image-LLMs to Long-Video-LLMs, aimed at improving long video understanding and reasoning. Following this, we review multiple video benchmarks of varying lengths and compare the video understanding performance of various methods. This comparison provides insights into future research directions for long video understanding. Our paper is the first to focus on the development and improvement of Long-Video-LLMs for better long video understanding. We hope our work will contribute to the advancement of long video understanding and reasoning with LLMs.

\section{Limitation}
We reviewed literature on comprehensive long video understanding, covering methods, training datasets, and benchmarks. Due to space constraints, we omit detailed application scenarios like real-time processing and multimodal tasks. We will maintain an open-source repository and add these contents to complement our survey. The performance comparisons are based on final results from previous papers and official benchmarks, which vary in training resources, strategies, and model architectures, making it difficult to analyze specific models and training differences. We plan to conduct detailed ablation studies on public benchmarks for a more direct analysis of model design, training resources, and methods.



\bibliography{main}

\appendix

\section{Example Appendix}
\label{sec:appendix}

This is an appendix.


\end{document}